\documentclass{article}
\usepackage{spconf,amsmath,epsfig}
\usepackage{url}
\usepackage{multirow}
\usepackage{hyperref}
\usepackage{graphicx, graphics}
\usepackage{xcolor}
\title{IMPROVED FLOOD MAPPING FOR EFFICIENT POLICY DESIGN BY FUSION OF SENTINEL-1, SENTINEL-2 AND LANDSAT-9 IMAGERY TO IDENTIFY POPULATION AND INFRASTRUCTURE EXPOSED TO FLOODS}
\name{U. Nazir$^{\star}$, M. A. Waseem$^{\star}$, F. S. Khan$^{\dagger}$, R. Saeed$^{\dagger}$, S. M. Hasan$^{\dagger}$, M. Uppal$^{\star}$, Z. Khalid$^{\star}$\thanks{We acknowledge the support of the Higher Education Commission of Pakistan under grant GCF-521.}}
\address{${\star}$Department of Electrical Engineering, Syed Babar Ali School of Science and Engineering\\
${\dagger}$ Department of Economics, Mushtaq Ahmad Gurmani School of Humanities and Social Sciences \\
Lahore University of Management Sciences (LUMS), Lahore, Pakistan\\
\{usman.nazir, m\_waseem, falak.khan, rabia.saeed, syed.hasan, momin.uppal, zubair.khalid\}@lums.edu.pk}
\begin{document}
%\ninept
%
\maketitle
\begin{abstract}
A reliable yet inexpensive tool for the estimation of flood water spread is conducive for efficient disaster management. The application of optical and SAR imagery in tandem provides a means of extended availability and enhanced reliability of flood mapping. We propose a  methodology to merge these two types of imagery into a common data space and demonstrate its use in the identification of affected populations and infrastructure for the $2022$ floods in Pakistan. The merging of optical and SAR data provides us with improved observations in cloud-prone regions; that is then used to gain additional insights into flood mapping applications. The use of open source datasets from WorldPop\footnote{\url{https://www.worldpop.org/}}and OSM\footnote{\url{https://www.openstreetmap.org/}} for population and roads respectively makes the exercise globally replicable. The integration of flood maps with spatial data on population and infrastructure facilitates informed policy design. We have shown that within the top five flood-affected districts in Sindh province, Pakistan, the affected population accounts for $31\%$, while the length of affected roads measures $1410.25$ km out of a total of $7537.96$ km.
\end{abstract}
\begin{figure*}[h]
    \centering
    \scalebox{0.82}{
    \begin{tabular}{ccccc}
      %\multirow{2}[2]{*}[20mm]{\includegraphics[scale=0.3, trim={0 0 0 0},clip]{images/FloodMappingv1.png}} & \includegraphics[scale=0.2, trim={0 0 3cm 0},clip]{images/FloodMappingv1.png} \\
     
      \includegraphics[scale=0.2, trim={11.5cm 0 11.5cm 0},clip]{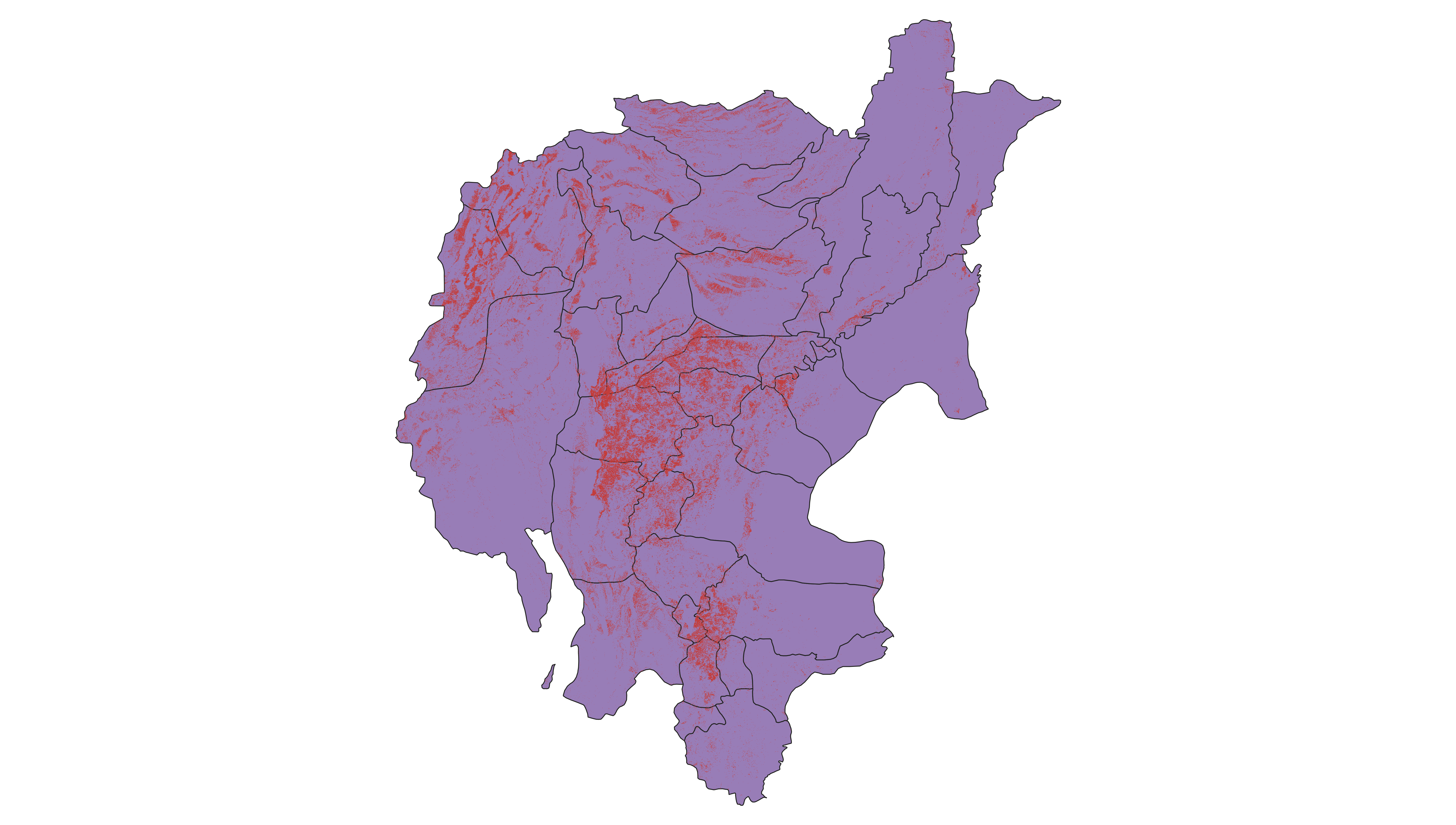} & \includegraphics[scale=0.2, trim={11.5cm 0cm 11.5cm 0}, clip]{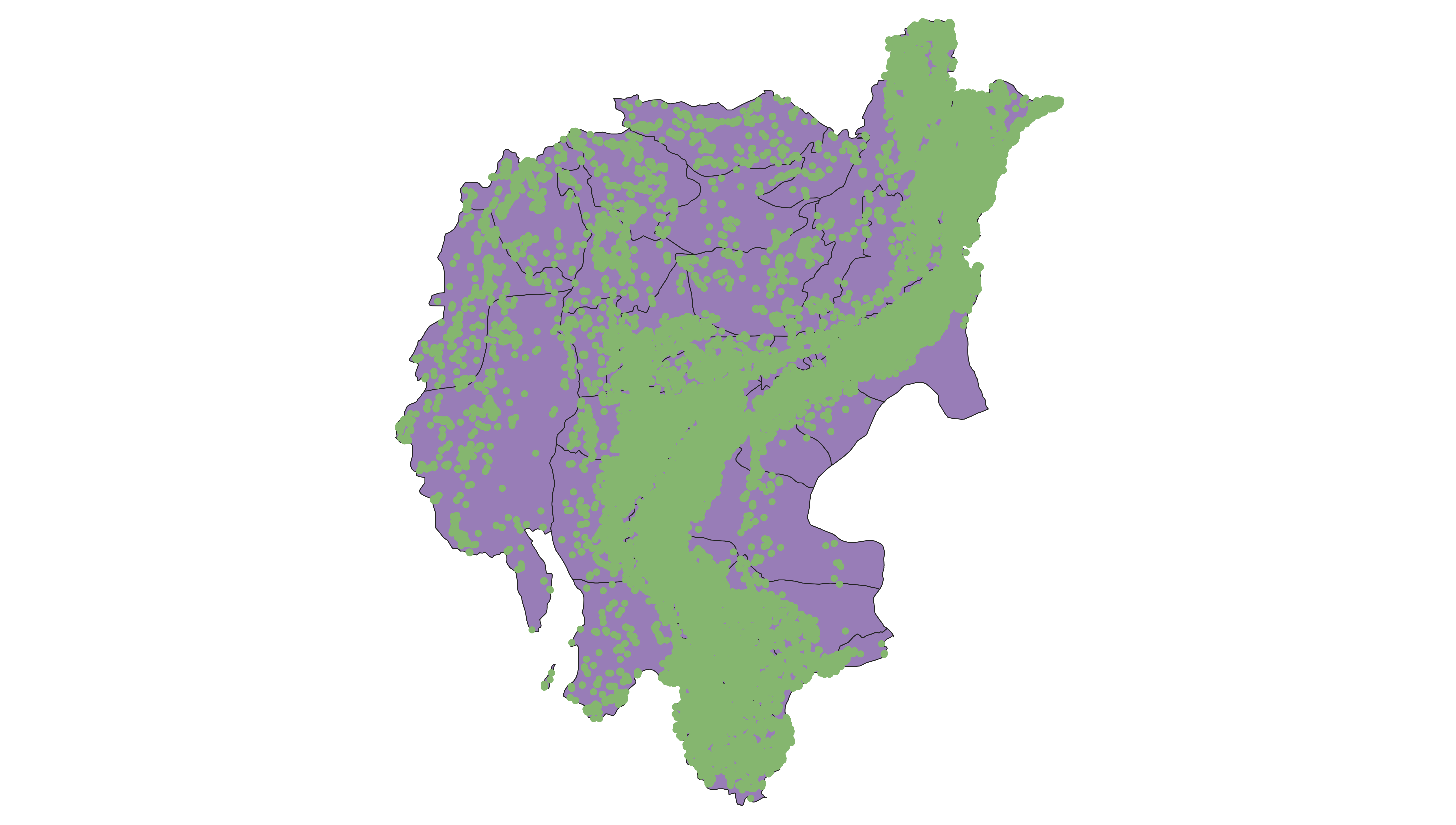} & \includegraphics[scale=0.2, trim={11.5cm 0 11.5cm 0},clip]{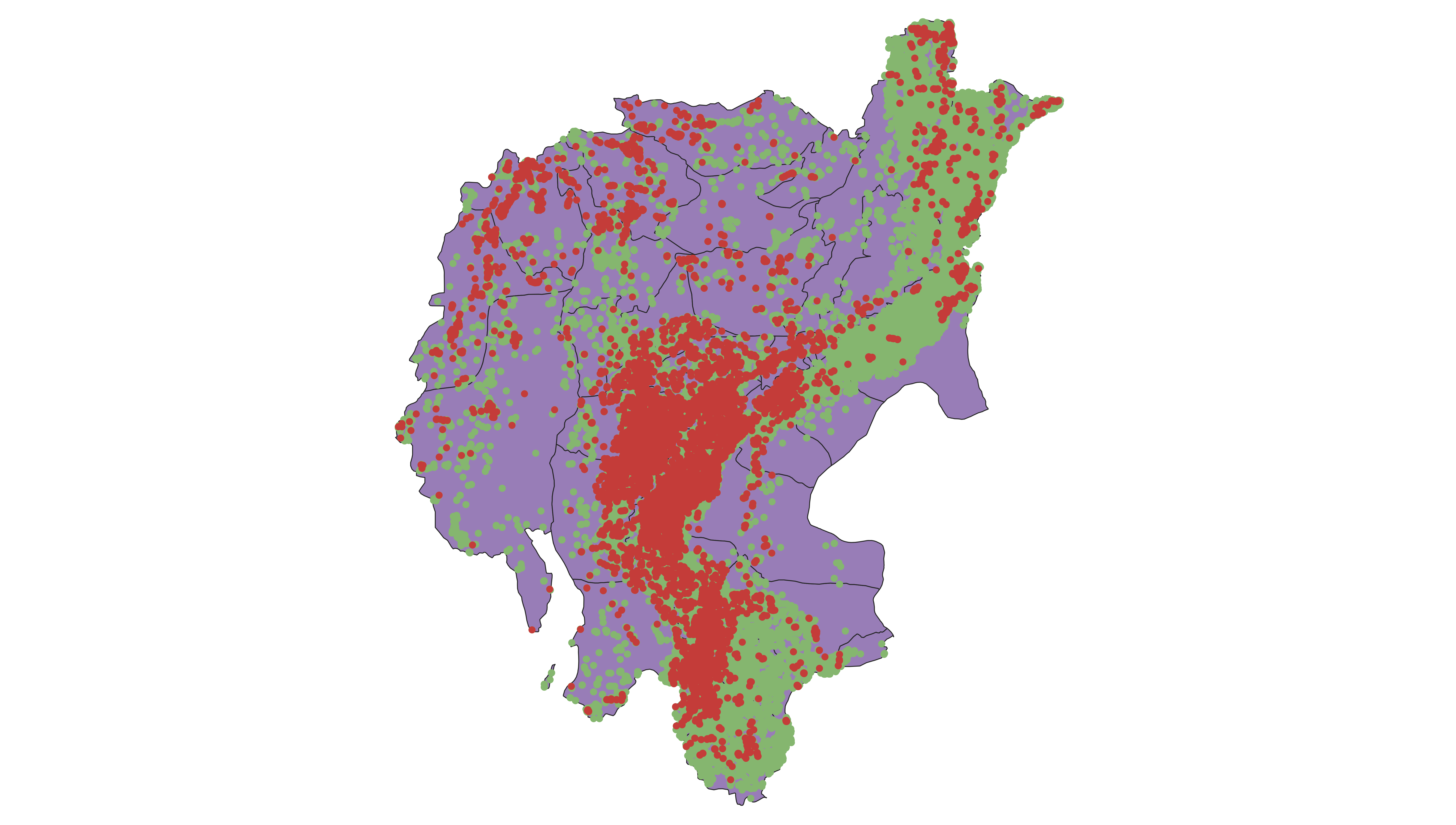} & \includegraphics[scale=0.2, trim={11.5cm 0 11.5cm 0}, clip]{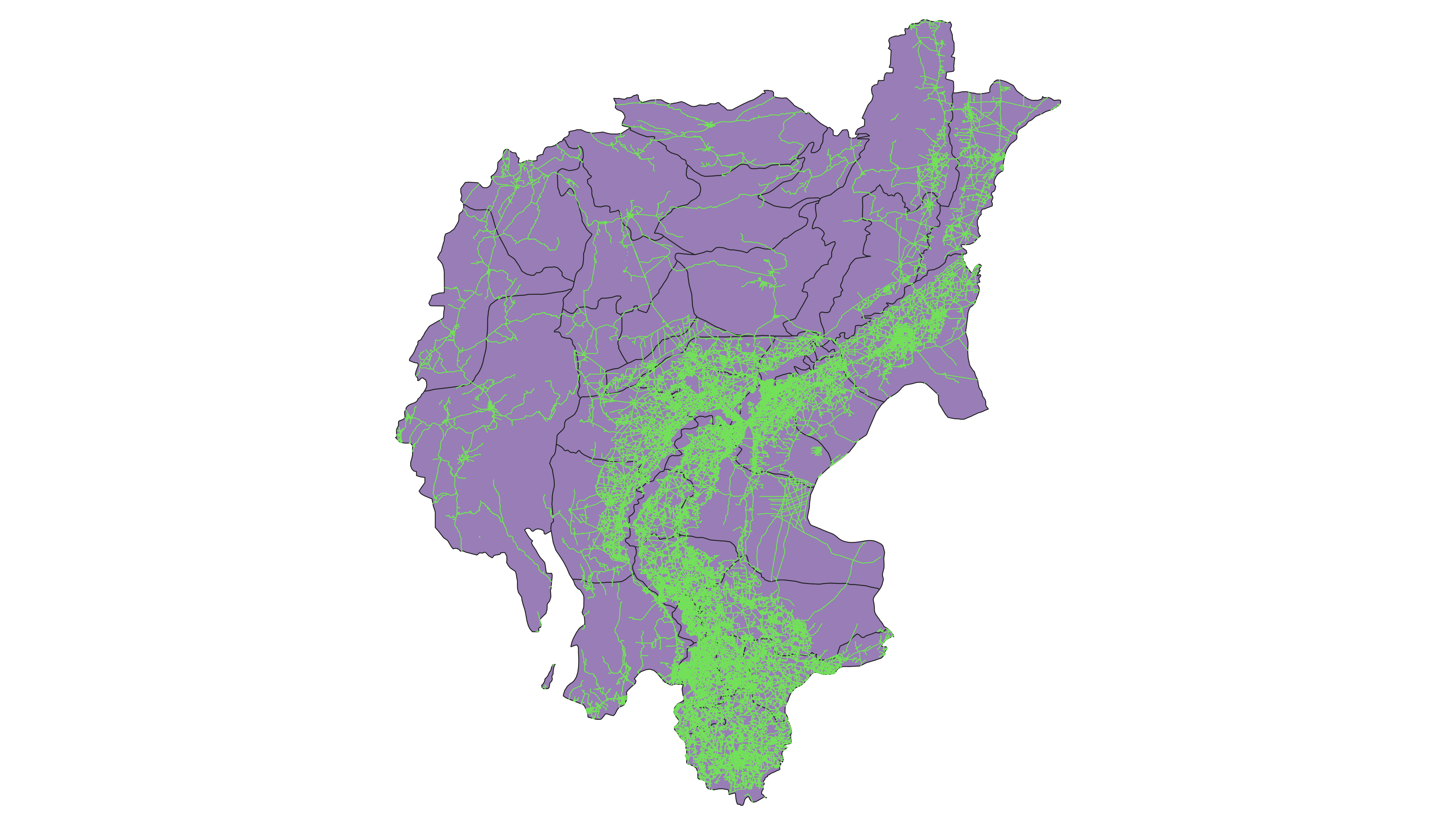}
      & \includegraphics[scale=0.2, trim={11.5cm 0 11.5cm 0},clip]{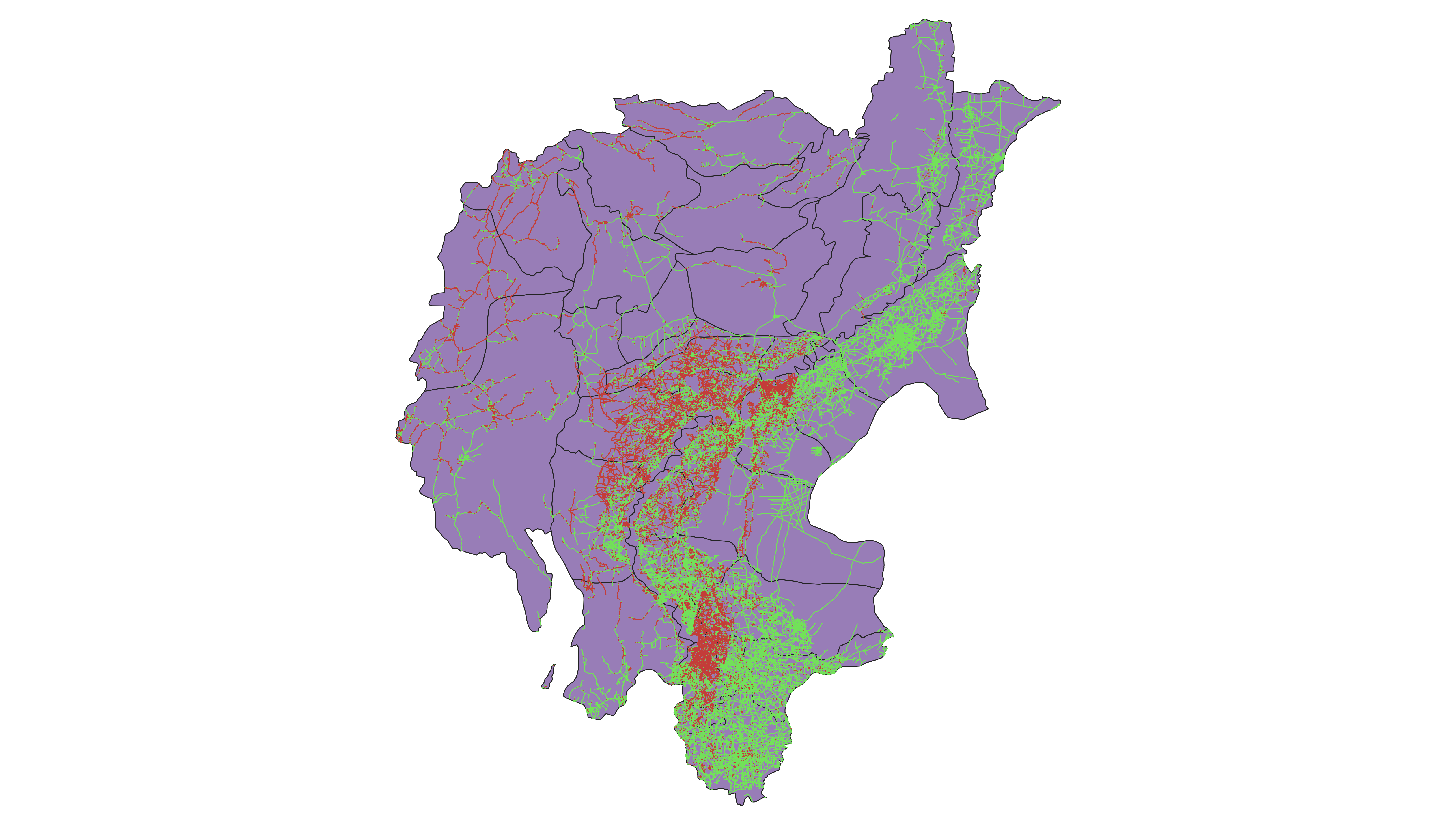} \\
      (a) & (b) & (c) & (d) & (e) 
    \end{tabular}}
    \caption{(a) Flood mapping in Sindh province, Pakistan using Sentinel-1, Sentinel-2 and Landsat-9 (radar and optical) satellites; (b) Population dataset from WorldPop; (c) Affected population (in red); (d) Roads network from OSM; (e) Affected roads (in red). }
    \label{fig:floodmappingworkflow}
\end{figure*}
\section{Introduction}
The availability of reliable data providing the spatial and temporal extent of a flooding event is a prerequisite for effective and efficient policy design and implementation. The damage, loss and needs assessment following a flood occurrence and the planning and execution of relief, rehabilitation and reconstruction measures are all contingent on flood information- period, extent, and depth of inundation.  Precipitation and inundation information collected through weather stations or aerial photography is often limited in time and space and hence needs to be supplemented with more periodic and wide-ranging satellite imagery. The policy relevance of this data is significant overall and besides its usefulness for relief and reconstruction, it can be helpful in the identification of disaster-prone infrastructure and assessing adverse impacts on the health and education outcomes of affected communities. The design of river flood defense requires the estimation of potential flood levels, extent, and period \cite{apel2004flood}. In the literature, a range of methodologies have focused on flood mapping using optical sensors (e.g. Landsat, Sentinel-2, and VIIRS)~\cite{huang2015evaluation, donchyts201630, pekel2016high} and SAR sensors (e.g. Radarsat and Sentinel-1)~\cite{clement2018multi, pham2017surface}. Flood mapping using optical sensors is hindered by clouds that obscure surface observations; resulting in data gaps. SAR data, on the other hand, can fill in gaps in optical data because SAR can penetrate cloud cover, operate in any weather conditions, and provide timely and crucial information about one of the most frequent and devastating natural disasters: flooding. Research has shown when flooding occurs, smooth water surfaces replace rough surfaces and reflect the radar signal in the specular direction at a distance from the antenna, resulting in a low back-scattering and showing as dark areas in SAR images~\cite{qiu2021flood}.

Few studies have explored the application of data fusion for flood mapping using both optical and SAR data~\cite{shakya2023fusion, quang2019synthetic, rudner2019multi3net, kwak2018improved}. In \cite{kwak2018improved}, statistical water index-based thresholding algorithm is used to detect and monitor mega river floods. In this paper, we proposed a novel tool for mapping the extent of floods. We utilized the Google Earth Engine, a cloud-based platform for processing remote sensing data. This platform offers enhanced computational speed as the processing is outsourced to Google's servers, eliminating the need to download raw imagery beforehand. The primary focus of the paper centers around highlighting the importance of integrating the generated flood maps with spatial data of population and infrastructure. Our proposed approach to flood mapping, will empower the decision-makers with the necessary information for effective disaster response and policy development.

%The proposed tool for flood extent mapping is expected to overcome the constraint of limited technical expertise that often hinders the availability of necessary information for effective disaster management. The use of Google Earth Engine for the cloud-based processing of remote sensing data is advantageous due to its enhanced computational speed, as processing is outsourced to Google’s servers and no prior download of raw imagery is required. The process is free of any charge, as only an active Google account with Google Earth Engine is required. Overlaying the flood maps with spatial data of population and infrastructure (roads, schools etc.) bridges the  gap for informed policy design. 
\section{Methodology for Mapping Floods}
To ensure adequate satellite image coverage for the desired area of interest, we establish specific pre- and post-flood time periods for radar (Sentinel-1) and optical (Sentinel-2 and Landsat-9) satellites.
\begin{table*}[h]
    \centering
    \caption{Quantitative analysis of population and infrastructure exposed to floods in Sindh province, Pakistan. Top-3 ranking flood affected districts are in bold and, in particular, red (1st), violet (2nd) and black (3rd).}
    \begin{tabular}{|c|c|c|c|c|}
     \hline    
      \textbf{Study area (districts)} & \textbf{Actual population} & \textbf{Affected population} & \textbf{Actual road length (km)} & \textbf{Affected road length (km)}  \\ \hline
      Jakobabad & 1382896 & \textcolor{red}{\bf 401370} & 3137.98 & \textcolor{red}{\bf 580.83} \\ \hline
      Mastung & 235830 & 112551 & 1102.04	& \textcolor{violet}{\bf 316.72} \\\hline
      Sibi & 233671 &	\textcolor{black}{\bf 114117} & 626.82 &	84.77 \\ \hline
      Jafarabad & 504570 & \textcolor{violet}{\bf 115279} & 926.50 & 117.93 \\ \hline
      Kalat & 290610 & 75822 & 1163.79 & \textcolor{black}{\bf 310} \\ \hline
    \end{tabular}
    \label{tab:quantitative}
\end{table*}
For optical-based satellites, flood mapping involves applying NDWI (Normalized Difference Water Index) thresholding to the difference layer between dry and wet images, utilizing water's spectral characteristics to identify flooded areas. Meanwhile, in radar-based satellites, the flood mapping process includes speckle filtering to remove noise, computing the difference between pre and post-flood mosaics, and applying post-processing filtering techniques to refine the results.

\begin{figure*}[h]
    \centering
    \scalebox{0.85}{
    \begin{tabular}{ccccc}     
      \includegraphics[scale=0.2, trim={11.5cm 0 11.5cm 0},clip]{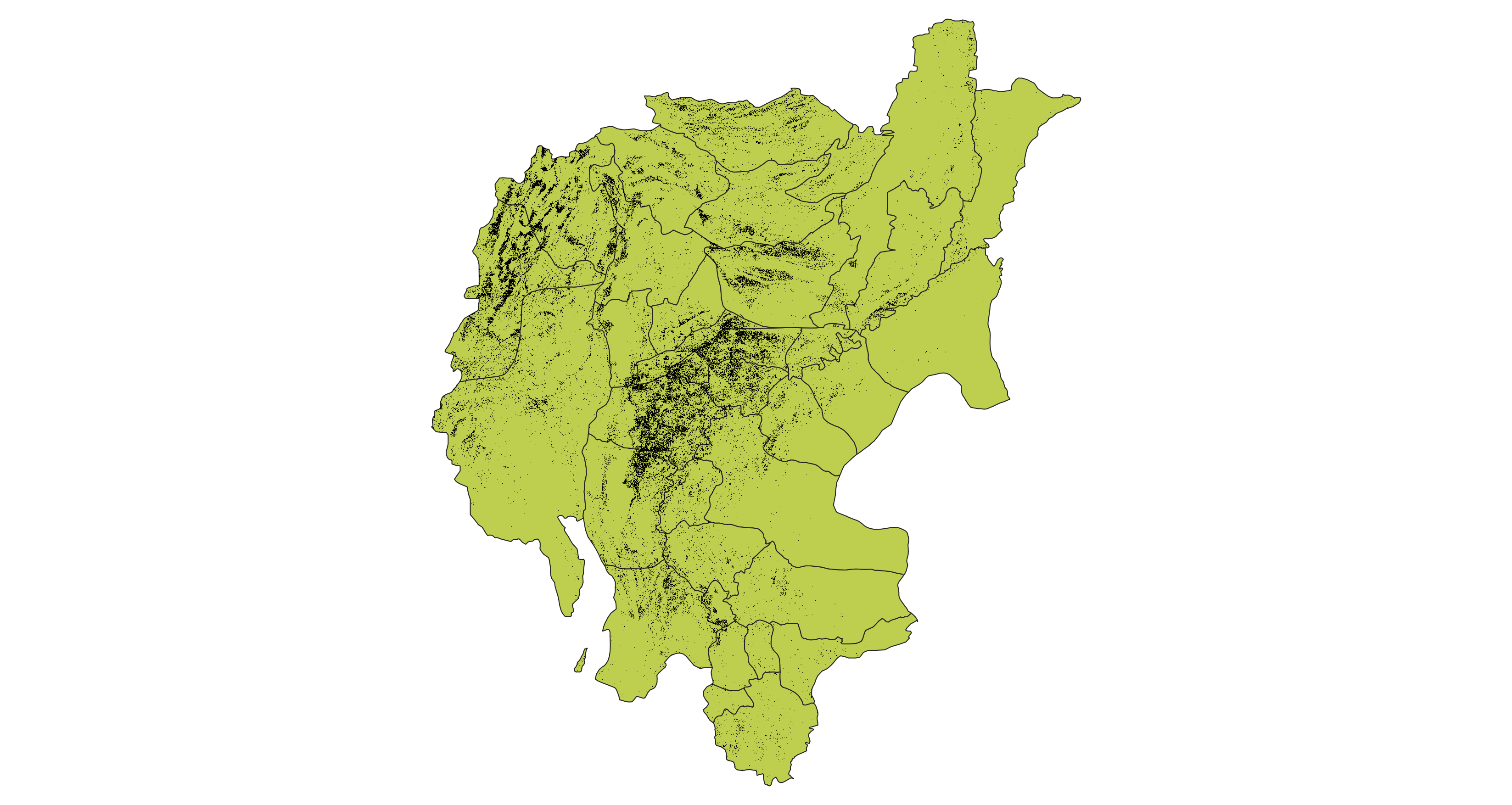} & \includegraphics[scale=0.2, trim={11.5cm 0cm 11.5cm 0}, clip]{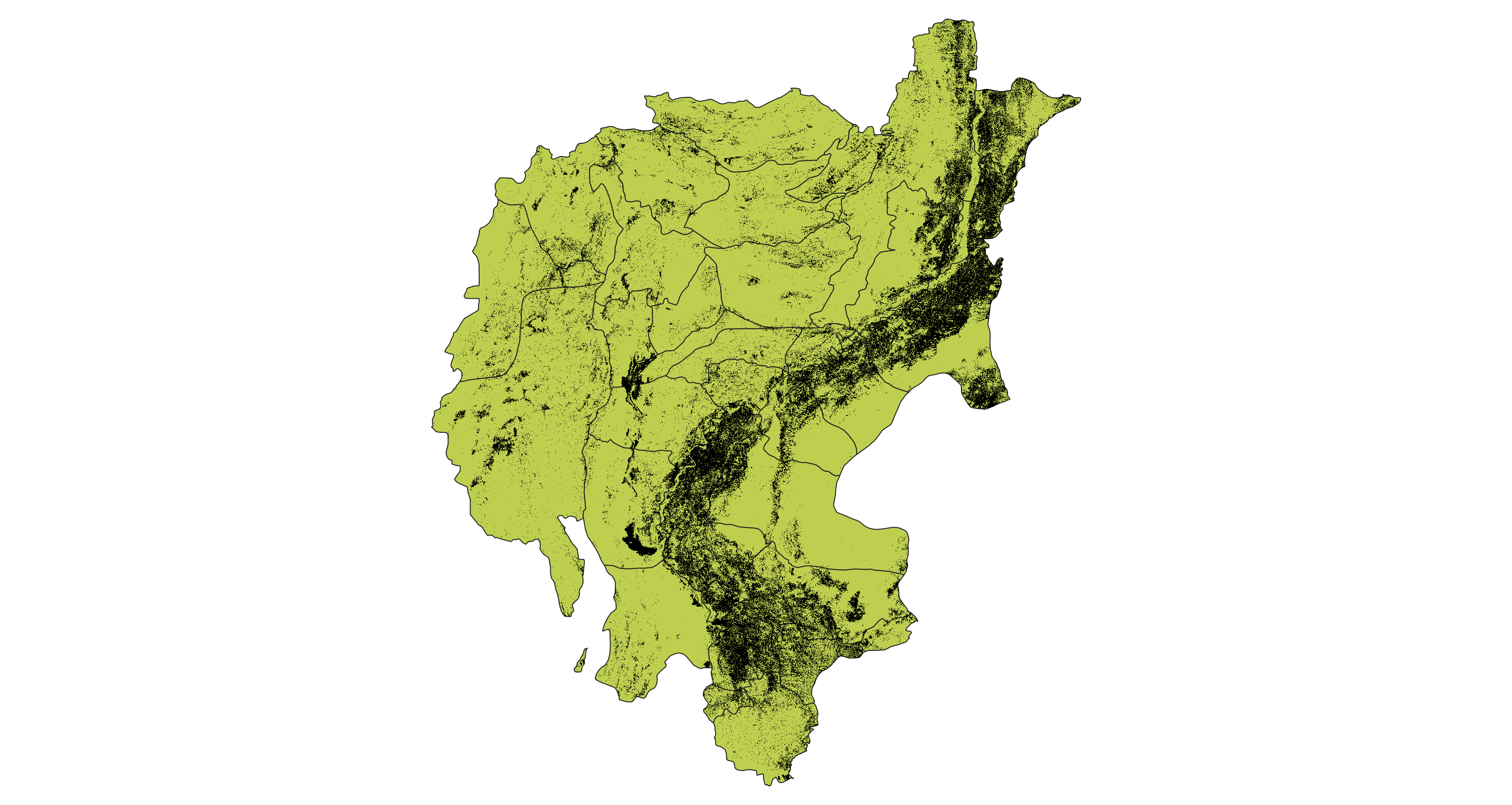} & \includegraphics[scale=0.2, trim={11.5cm 0 11.5cm 0},clip]{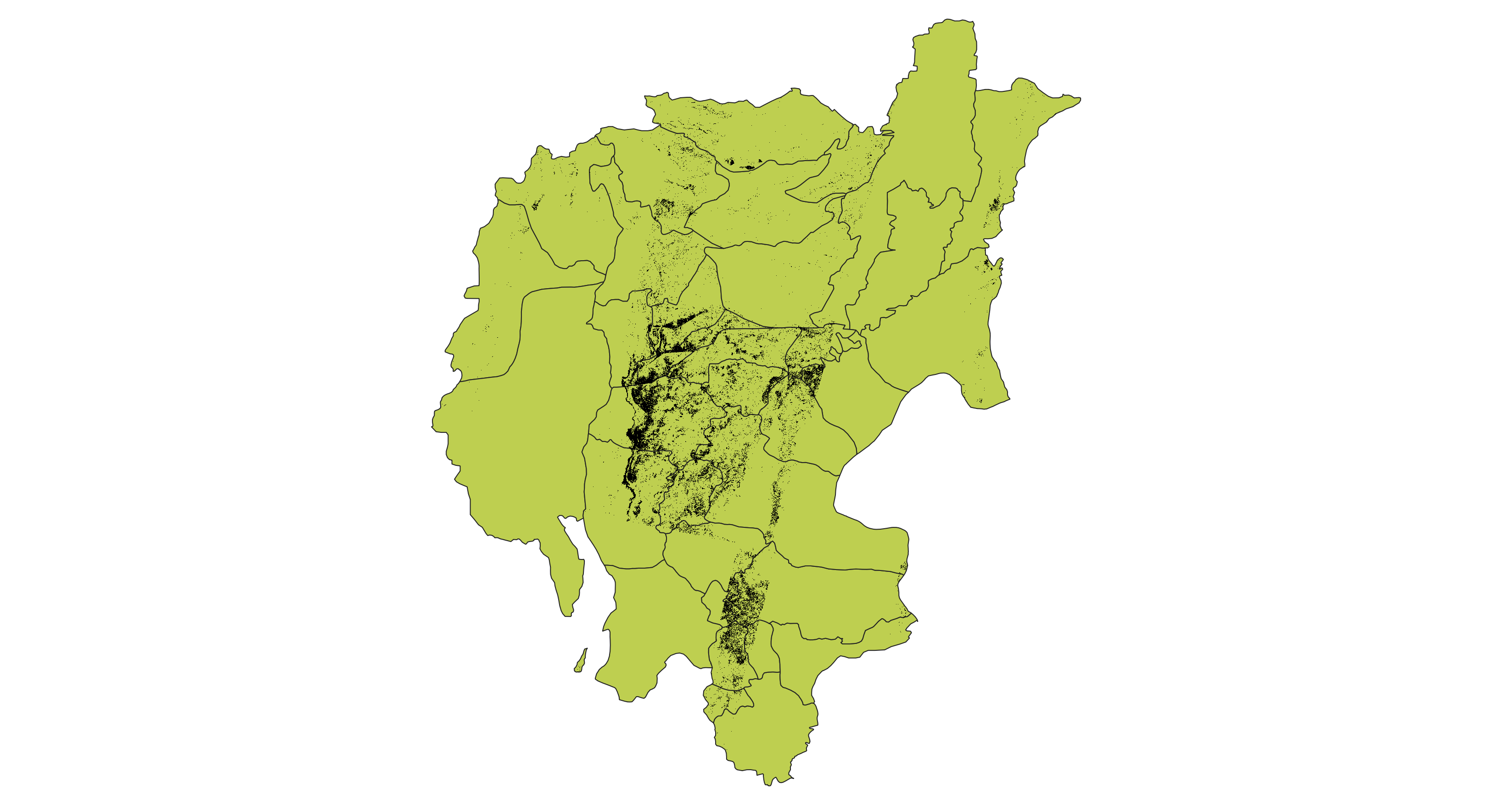} & \includegraphics[scale=0.2, trim={11.5cm 0 11.5cm 0}, clip]{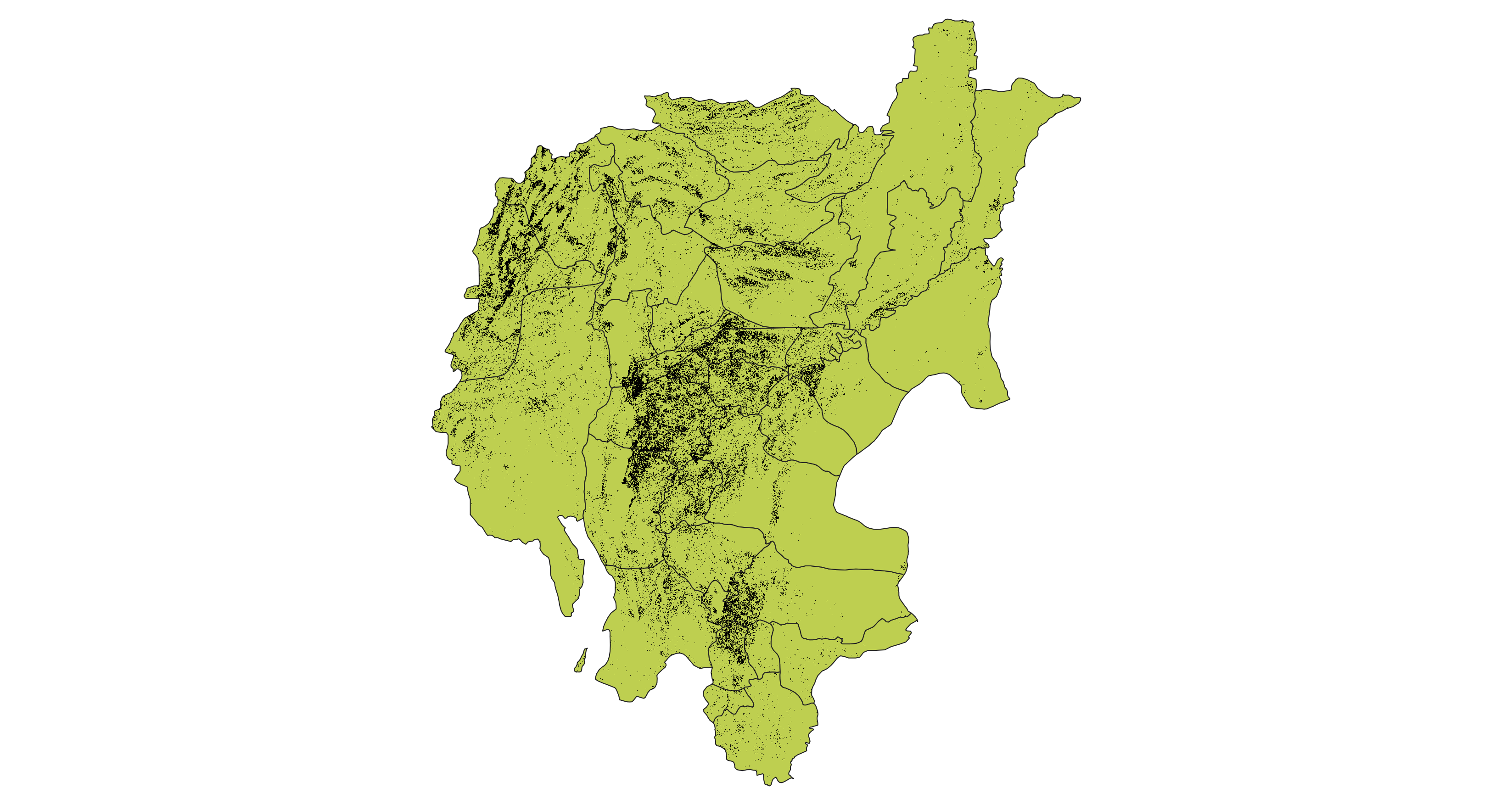} &
      \includegraphics[scale=0.35, trim={0 1.5cm 15cm 0}, clip]{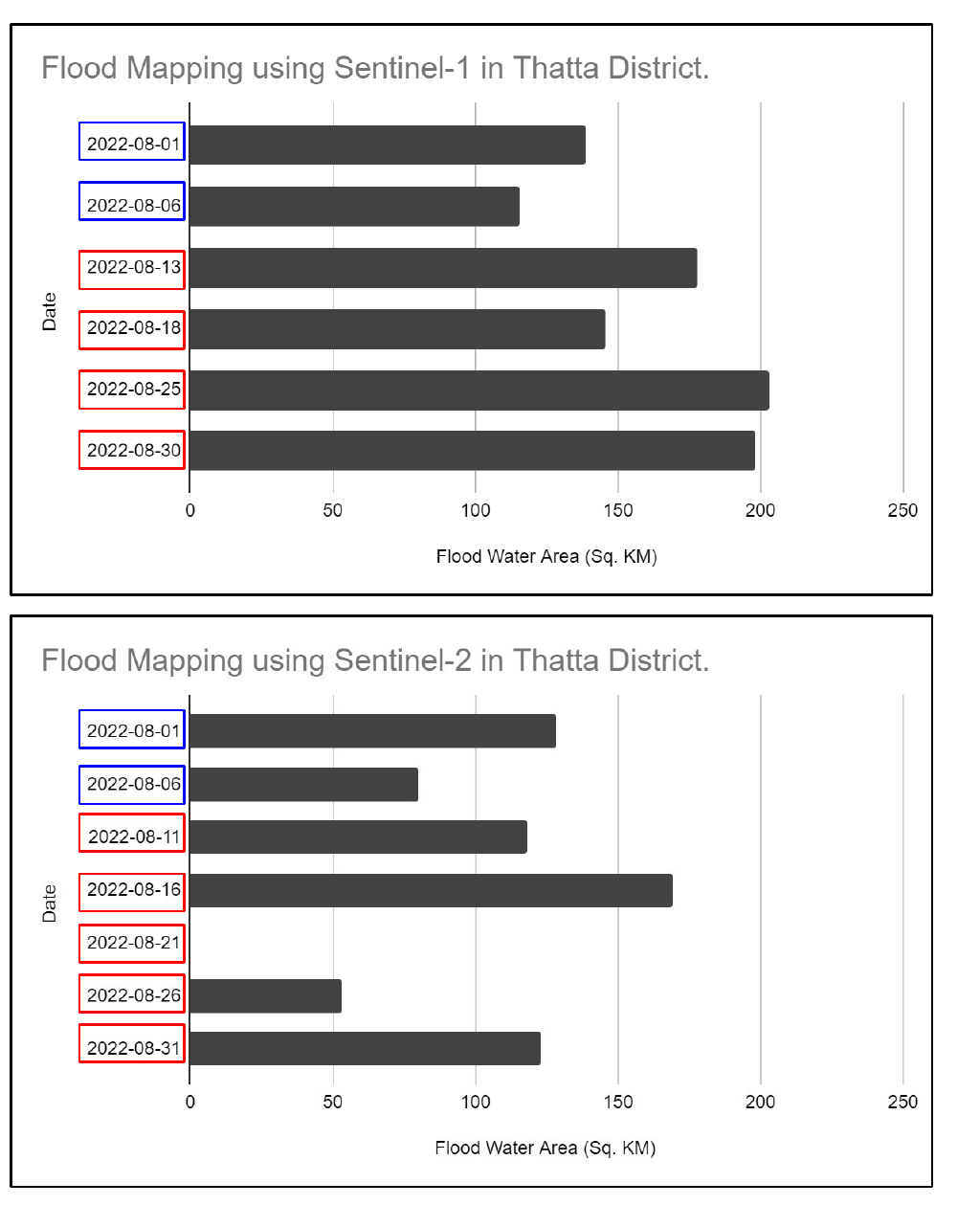} \\
      (a) & (b) & (c) & (d) & (e) 
    \end{tabular}}
    \caption{\emph{Flood mapping} in Sindh province, Pakistan using (a) Sentinel-1, (b) Sentinel-2, (c) Landsat-9,  (d) Intersection and majority voting; (e) Availability of flood mapping (continuous monitoring) is increased using multiple satellites for the month of August 2022 with union operation as Sentinel-1 and Sentinel-2 images are available for different dates (in red lines). }
    \label{fig:floodmapping}
\end{figure*}

To enhance the reliability of flood mapping, a fusion approach is employed, where the intersection of the flood mapping results from optical satellites and the flood map generated from Sentinel-1 radar data is taken on a pixel-wise basis. This fusion process leverages the strengths of both optical and radar-based observations, resulting in a more accurate and comprehensive flood mapping output. (see Fig.~\ref{fig:floodmapping}). 

\section{Evaluation Results}
\subsection{Study Region}
For the purpose of identifying the population and roads vulnerable to flooding, we choose the entire province of Sindh, Pakistan as our study area and make use of open source datasets provided by WorldPop and OpenStreetMap (OSM) to obtain population and road information respectively. 

To identify schools impacted by floods, we chose seven districts in the Sindh province of Pakistan as our study area: Hyderabad, Jamshoro, Malir, Malir Cantonment, Sajawal, Thatta, and T. M Khan (See Fig.~\ref{fig:schools}).

\subsection{Datasets}
\subsubsection{Population Data}
WorldPop is a research group focusing to the provision of open access spatial demographic datasets. They utilize remote sensing and geospatial analysis, to estimate and map population distributions at fine scales across the globe. The datasets offered by WorldPop have diverse applications, including urban planning and disaster management.
\subsubsection{Roads Network}
The roads network used is sourced from OpenStreetMap (OSM). OSM is an open collaborative mapping project that provides a wealth of geographic information, including roads, buildings, landmarks, and more.
\subsubsection{School Locations}
To determine the geo-locations (geo-coordinates) of the schools in the study area, we utilize a classical computer vision method, namely `Template Matching'. First, we download high resolution (zoom level 21\footnote{At zoom level 21, pixel resolution is 0.075 meters.}) google map imagery for our region of interest. Google maps provide precise locations of different points of interest (POIs) (education, food, park, health, etc.), each of which can be separated using a unique icon. So, we simply extract the icon-image of education POIs, and we apply template matching between this icon-image and the google map images. For template matching, we use the normalized correlation coefficient method and find the locations where this correlation coefficient value is greater than $0.95$. Since the template matching is applied on simple images, we obtain pixel locations after applying template matching. Hence, we interpolate the geo-spatial extents of the google map images to convert these pixel locations into geo-coordinates. We also add a post-processing step to ensure that all the extracted points are at least $10$ meters apart, as it is practically impossible to have more than one educational building in such a small radius. Using template matching with Google Maps images provides a practical approach to efficiently identify school buildings (see Fig.~\ref{fig:schools}(a)).
\subsection{Affected Population and Roads}
To identify the population and infrastructure exposed to floods, we take the pixel wise intersection of final flood map (see Fig.~\ref{fig:floodmappingworkflow}(a) and Fig.~\ref{fig:floodmapping}(d)) with the population map from WorldPop (see Fig.~\ref{fig:floodmappingworkflow}(b)) and the roads network from OSM (see Fig.~\ref{fig:floodmappingworkflow}(d)) respectively. We get the affected population in  Fig.~\ref{fig:floodmappingworkflow}(c) and affected roads in Fig.~\ref{fig:floodmappingworkflow}(e). We also achieve increased availability of flood mapping by computing the flood areas in different  districts using different satellites as optical and radar based satellites provide imagery for different dates (see Fig.~\ref{fig:floodmapping}(e)).
Table~\ref{tab:quantitative}, shows the estimated number of the affected population and the length of damaged roads (in KM) in \emph{top-$5$} flood affected districts (out of $35$) of Sindh province, Pakistan.

\begin{table}[h]
    \centering
    \caption{Affected schools in study area of Sindh province. Top-3 ranking flood affected districts are in bold and, in particular, red (1st), violet (2nd) and black (3rd).}
    \begin{tabular}{|c|c|c|}
        \hline
        \textbf{Study area (districts)} & \textbf{Total schools} & \textbf{Affected schools}  \\ \hline
        Jamshoro & 366 & \textcolor{black}{\bf 5} \\ \hline
        Malir & 1242 & \textcolor{red}{\bf 7}\\ \hline
        Malir Cantonment & 88 & 0\\ \hline
        Thatta & 384& \textcolor{violet}{\bf 6}\\ \hline
        Sujjawal & 246 & 1\\ \hline
        Hyderabad & 625 & \textcolor{black}{\bf 5} \\ \hline
        T. M Khan & 33 & 0 \\ \hline
        
    \end{tabular}
    \label{tab:Aschools}
\end{table}
%\subsection{Affected Educational Institutes}
\subsection{Affected Schools}
\begin{figure}[h]
    \centering
\scalebox{0.6}{
    \begin{tabular}{ccc}
    \includegraphics[scale = 0.45, trim={10cm 2cm 10cm 2cm},clip]{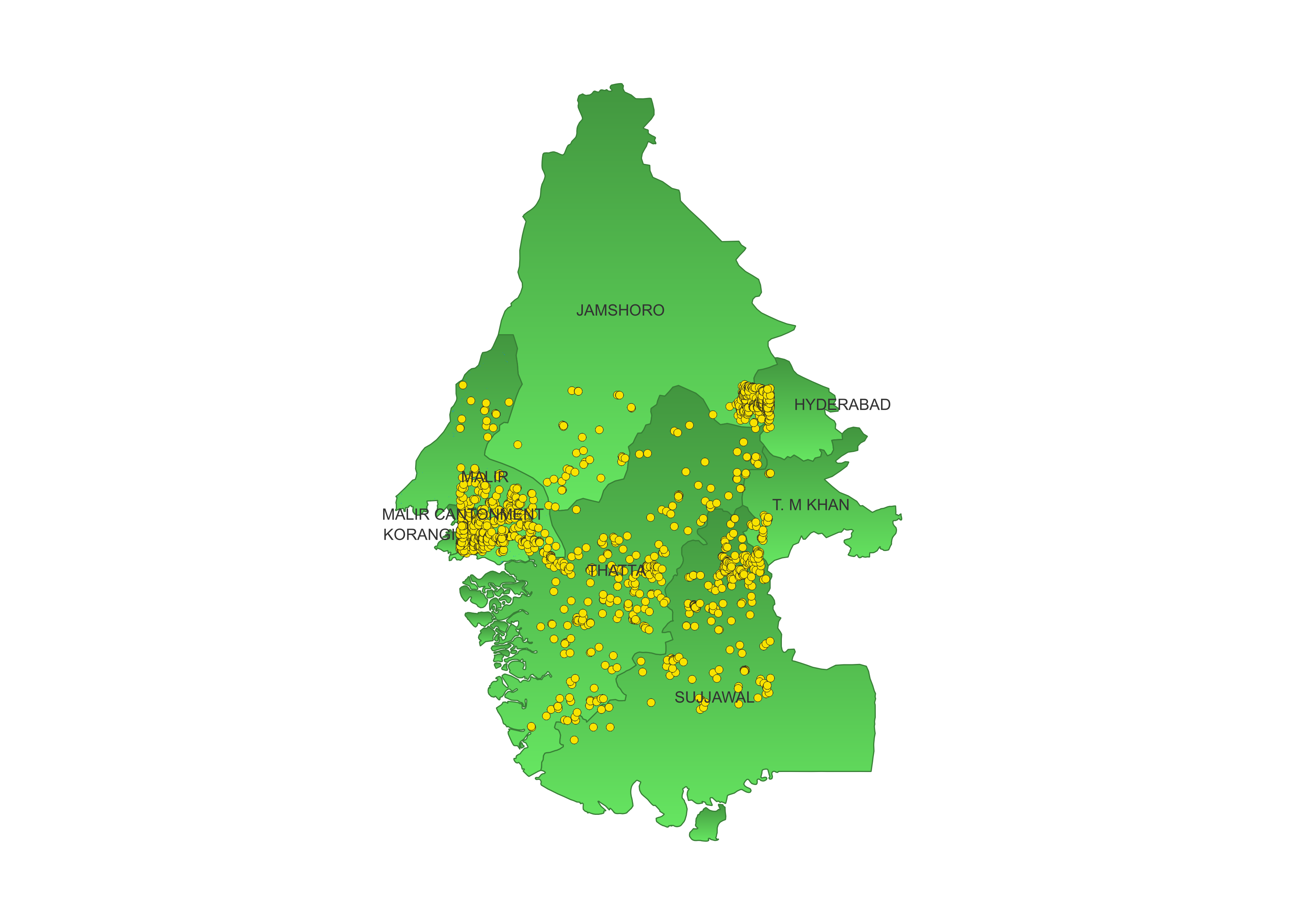} & \includegraphics[scale = 0.45, trim={10cm 2cm 10cm 2cm},clip]{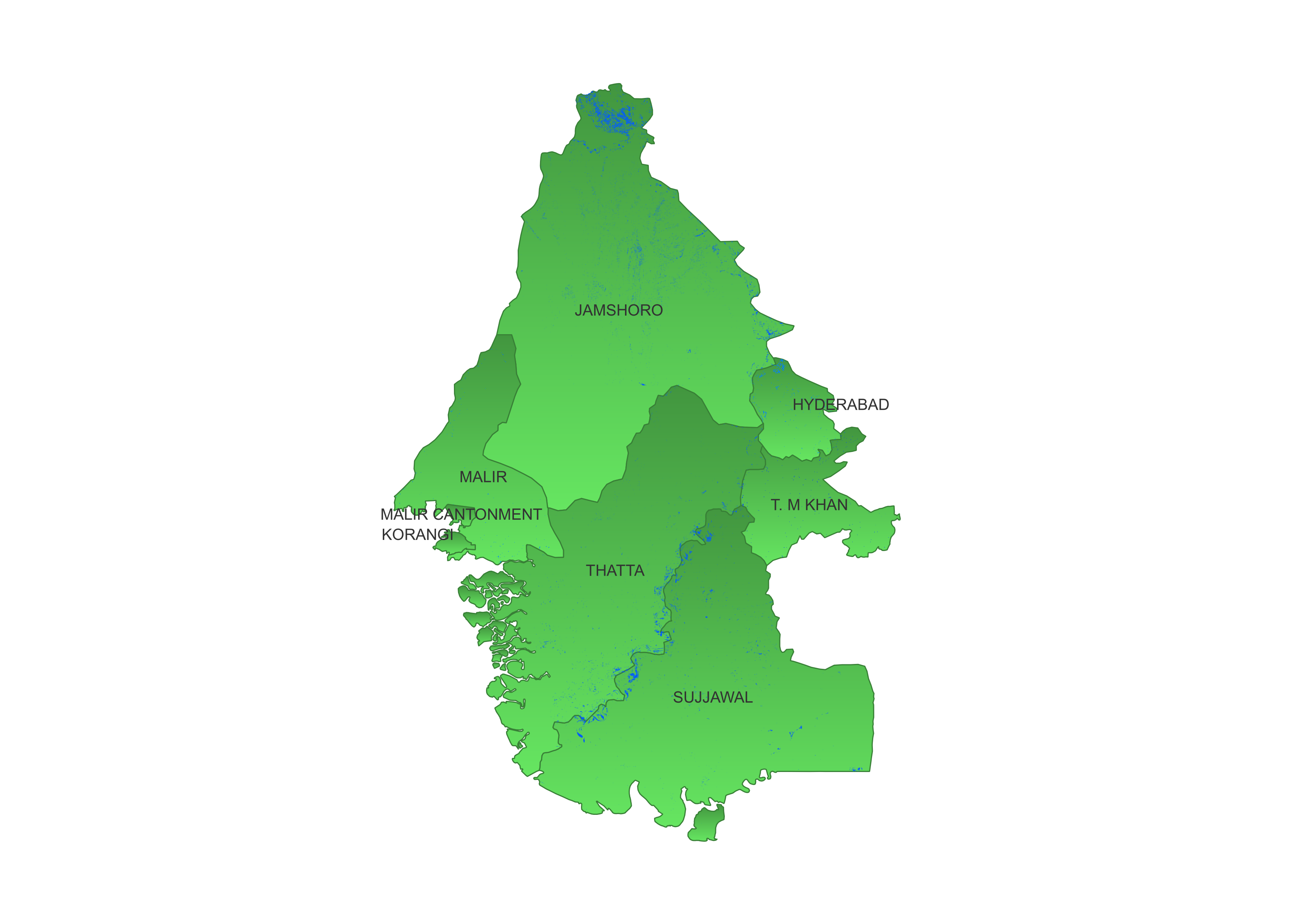} & \includegraphics[scale = 0.45, trim={10cm 2cm 10cm 2cm},clip]{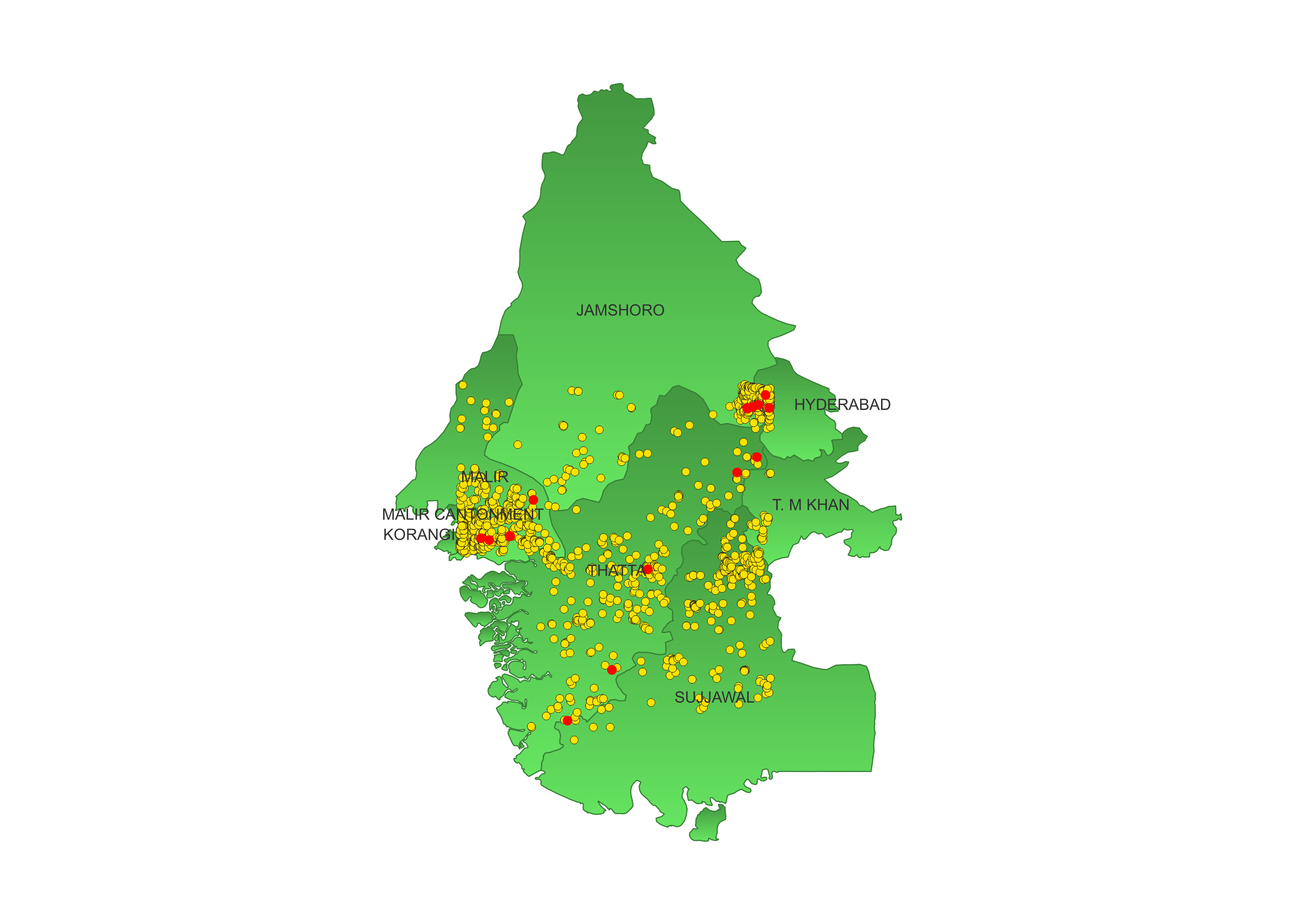}\\
    (a) & (b) & (c) \\
    \end{tabular}}
    \caption{(a) The study area designated as the green region encompasses the locations of schools, which are indicated by yellow dots, (b) Flood mapping (blue dots) in the study region, (c) Affected schools (red dots).}
    \label{fig:schools}
\end{figure}

We employed the intersection of school locations and vectorized flood maps to identify the affected schools in our study region. By comparing the location data from both maps, we identified $24$ schools that have been impacted. These affected schools are depicted as red dots in Fig.~\ref{fig:schools}(c). Table~\ref{tab:Aschools} shows that the highest number of flood-affected schools can be observed in Malir and Thatta districts.
\section{Flooding and Education Outcomes: Present Situation and Updates}
%\subsection{Current Status of Education Outcomes}
A field survey conducted from $24$th to $27$th November 2022, assessed the flooding levels and impacts on education in various locations in Sindh, Pakistan. The survey covered eight areas: Hyderabad, Jamshoro, Matyari, Sehwan, Manchar lake, Mehar, Nawabshah, and Bhit Shah. While it was not feasible to conduct full-length surveys to assess learning losses in the affected districts given the time and budget constraints, we plan to do the same in the future for $ 35 $ districts in Sindh province. Nonetheless, we used nationally representative data from Pakistan Social and Living Standards Measurement Survey (2018-19 round) to develop a sense of existing disparities in the affected districts. We find that the districts that are overall affected by the incidence of flood  i.e. Jacobabad and Sibi, as well as the ones in which we observe the maximum number of schools affected i.e. Thatta and Malir; already lag behind in certain literacy and education outcomes. For instance, the literacy rate (those who can read, and write and solve basic numeracy questions) is $23 $  \% compared to $ 38 $ \% in other districts of the same province. Similarly, in the districts where most schools are affected, $64$ \% population has never attended school compared to $ 43.7 $ \% in other districts. 
%\subsection{Status Update on Flooding}

The survey revealed three types of flooding scenarios. Firstly, there was an increase in surface water flowing from the mountainous regions of Balochistan towards the flat fields of Sindh. Secondly, the Indus river overflowed into the floodplain. Lastly, prolonged intense rainfall on flat terrain resulted in flooding. The water levels varied across these regions. In the croplands within the floodplain, the water has mostly dried out naturally or has been pumped out. However, in areas like Mehar and beyond, the water levels still remain around 2 feet deep, causing displacement of people and hindering the possibility of sowing crops.

\section{Conclusion} The proposed methodology for flood mapping has certain advantages: (a) useful in spatially identifying  flood-affected households and disaster-prone infrastructure such as roads, schools etc. to aid in informed and data driven policy design, (b) uses publicly available satellite imagery and cloud-based processing (c) requires no specific hardware. Our proposed methodology will help us target areas for surveys in a reliable fashion to compare education (and more socioeconomic) outcomes in areas differently affected by floods. 

%\pagebreak

\bibliographystyle{IEEEbib}
\bibliography{refs}

\end{document}